%% file: AnonymousSubmission2027.tex
\documentclass[letterpaper]{article} 
\usepackage[preprint]{aaai2027}  
\usepackage[hyphens]{url}  
\usepackage{graphicx} 
\usepackage{natbib}  
\usepackage{caption} 
\usepackage{amsmath}
\usepackage{amsfonts}
\usepackage{amssymb}
\usepackage{bm}
\usepackage{booktabs}
\usepackage{algorithm}
\usepackage{algorithmicx}
\usepackage{algpseudocode}
\usepackage{enumitem}
\usepackage{newfloat}
\usepackage{listings}

\definecolor{PromptInk}{HTML}{315873}
\definecolor{PromptRule}{HTML}{91A6B5}
\definecolor{PromptFill}{HTML}{F4F7F9}

\lstdefinestyle{promptbox}{%
  basicstyle=\fontsize{6.7}{7.3}\selectfont\ttfamily,
  numbers=none,
  xleftmargin=1pt,
  xrightmargin=1pt,
  framexleftmargin=4pt,
  framexrightmargin=4pt,
  framextopmargin=3pt,
  framexbottommargin=3pt,
  aboveskip=3pt,
  belowskip=3pt,
  showstringspaces=false,
  columns=fullflexible,
  keepspaces=true,
  tabsize=2,
  breaklines=true,
  breakindent=0pt,
  breakautoindent=false,
  frame=single,
  framerule=0.6pt,
  framesep=3pt,
  rulecolor=\color{PromptRule},
  backgroundcolor=\color{PromptFill},
  escapeinside={(*@}{@*)}}

\DeclareCaptionStyle{ruled}{labelfont=normalfont,labelsep=colon,strut=off} 
\floatstyle{ruled}
\newfloat{listing}{tb}{lst}{}
\floatname{listing}{Listing}

\title{ScaFE: Data-Efficient Scar Classification with LLM-Generated Clinical Feature Programs}

\author{
    Ruman Wang\textsuperscript{\rm 1}, Hangting Ye\textsuperscript{\rm 2},
}
\affiliations{
    \textsuperscript{\rm 1}Liaoning University of Traditional Chinese Medicine; \textsuperscript{\rm 2}School of Artificial Intelligence, Jilin University\\

    ruman.wang@outlook.com, yeht22@mails.jlu.edu.cn
}

\begin{document}

\maketitle

\input{sections/abstract}

\input{sections/introduction}

\input{sections/related-work}

\input{sections/preliminary}

\input{sections/method}

\input{sections/experiment}

\input{sections/conclusion}

\bibliography{aaai2027}





\end{document}

%% file: sections/abstract.tex
\begin{abstract}
Classifying pathological scars from clinical photographs requires distinguishing keloids from hypertrophic scars despite limited expert-labeled data and substantial acquisition variation across hospitals. End-to-end image models remain data-dependent, whereas sending photographs to a hosted vision-language model (VLM) may conflict with local data-governance requirements and yields decisions that are difficult to reproduce and audit. We introduce \textbf{ScaFE} (\textbf{Sc}ar \textbf{F}eature \textbf{E}ngineering), which transfers clinical knowledge from a large language model (LLM) into deterministic, executable feature programs instead of asking the model to diagnose images. A web-enabled LLM retrieves clinical evidence and synthesizes programs that measure visually assessable scar attributes. Candidate programs execute in a restricted local environment, and only aggregate validation statistics and feature-level SHAP summaries are returned for iterative repair and refinement; raw images and patient-level outputs remain local. A lightweight Random Forest then operates on the resulting structured representation. On 600 photographs from three hospitals under leave-one-site-out evaluation, ScaFE achieves 81.0\% site-macro balanced accuracy, exceeding the strongest baseline, BiomedCLIP, by 10.0 percentage points. With only 10\% of the development data, ScaFE retains 72.0\% balanced accuracy and an 11.8-point lead. Refinement raises executability from 66.7\% to 95.0\% and the candidate evidence-pass rate from 70.0\% to 91.7\%, while final filtering ensures 100\% coverage among retained features. These results show that LLM knowledge can support data-efficient, cross-site medical image classification through local and auditable feature programs rather than direct VLM decisions.
\end{abstract}

%% file: sections/introduction.tex
\section{Introduction}

Keloids (KD) and hypertrophic scars (HS) are pathological responses to wound healing that can appear similar in clinical photographs but differ in growth pattern, prognosis, and treatment. Keloids extend beyond the original wound and commonly recur, whereas hypertrophic scars remain within the wound boundary and may regress over time~\cite{bayat2003skin,berman2017keloids}. Distinguishing them affects treatment planning. Because clinical scar assessment relies on observer-scored properties, the original VSS and POSAS studies explicitly evaluate interrater reliability~\cite{baryza1995vancouver,draaijers2004posas}. Reproducible image-based decision support could therefore provide a standardized second opinion or triage aid when scar expertise is not immediately available; it is intended to assist, not replace, clinical evaluation.

Building such a system presents two coupled challenges. \textbf{First, scar representations must be data-efficient and generalize across clinical sites.} Expert annotation is costly, and specialized scar cohorts are modest relative to the data typically used to learn image representations. Photographs also vary across hospitals in camera, illumination, viewpoint, skin tone, and anatomical site. Limited labeled cohorts and heterogeneous acquisition are persistent challenges in medical image learning~\cite{litjens2017survey}. The relevant difficulty is thus not an inability to collect data from multiple hospitals---our cohort spans three---but learning a stable decision rule from a limited number of specialist-labeled cases.

\textbf{Second, clinical knowledge in LLMs must be externalized without turning the LLM into an opaque image classifier.} LLMs have demonstrated medical question answering, knowledge recall, and reasoning capabilities~\cite{singhal2023large,nori2023capabilities}. Direct multimodal inference, however, either sends photographs to a hosted service or requires a large model to be deployed locally. In both cases, the prediction remains tied to a black-box, potentially variable generation process; a pilot study of chat-based models on KD--HS images also found that direct diagnosis was not yet clinically adequate~\cite{shiraishi2024potential}. Generating code avoids direct diagnosis, but introduces its own reliability problem: a plausible program may fail to execute, measure the wrong visual property, or lack clinical support. A useful knowledge-transfer mechanism must therefore keep images local while producing executable, evidence-grounded, and auditable measurements.

These challenges lead to our central question: \emph{Can an LLM contribute clinical knowledge to data-efficient medical image classification without observing patient images or making the final diagnosis?} Scar assessment offers a natural interface between textual knowledge and image computation. Instruments such as the Vancouver Scar Scale (VSS) and Patient and Observer Scar Assessment Scale (POSAS) organize scar evaluation around named properties, including pigmentation and vascularity~\cite{baryza1995vancouver,draaijers2004posas}. Although photographs cannot capture scale items that require palpation or patient report, their visually assessable concepts suggest a structured representation: convert observable image patterns into explicit measurements, then learn a decision rule in that low-dimensional space.

Building on this insight, we propose \textbf{ScaFE} (\textbf{Sca}r \textbf{F}eature \textbf{E}ngineering), a bounded program-synthesis framework that uses an LLM as a knowledge-driven feature engineer. A web-enabled LLM first retrieves clinical sources under a traceability contract, then generates candidate Python programs whose output dimensions have names, operational definitions, clinical rationales, and supporting evidence. The programs run locally under a restricted interface with no access to the network, labels, filenames, metadata, or patient records. A fixed Random Forest (RF) evaluates each representation on an inner validation split. ScaFE returns only aggregate execution errors, class-wise confusion counts, balanced accuracy, invalid or constant feature rates, and global SHAP importance to the LLM. Over several rounds, this feedback guides code repair and feature revision without exposing any image or patient-level output. Unsupported dimensions are removed by a final evidence gate, and the selected program and RF are frozen before evaluation on an unseen hospital.

This design separates three roles that end-to-end and direct-VLM approaches conflate: the LLM supplies clinical priors, deterministic code performs local measurement, and a lightweight learner estimates the diagnostic boundary. The separation reduces dependence on labeled images, makes each feature inspectable, and permits a hosted LLM to assist without receiving clinical photographs. In leave-one-site-out evaluation on 600 images from three hospitals, ScaFE reaches 81.0\% site-macro balanced accuracy, 10.0 points above the strongest baseline. Its advantage increases to 11.8 points when only 10\% of the development cohort is available, and ablations verify the contributions of literature grounding and semantically aligned validation feedback.

Our contributions are threefold:
\begin{itemize}[leftmargin=*,itemsep=2pt,topsep=2pt]
    \item We formulate LLM-assisted medical image learning as evidence-grounded feature-program search and introduce ScaFE, which transfers clinical knowledge into deterministic local measurements rather than direct image predictions.
    \item We develop a validation-guided refinement loop that improves program executability and feature utility using only aggregate feedback, while retaining a traceable evidence record for every feature used by the final predictor.
    \item We conduct patient-level leave-one-site-out evaluation across three hospitals, demonstrate gains over supervised, foundation-model, handcrafted, and direct-VLM baselines, and audit executability, evidence grounding, feature faithfulness, and cross-LLM stability.
\end{itemize}

%% file: sections/related-work.tex
\section{Related Work}

\subsection{Scar Assessment and Image-Based Learning}
Clinical scar assessment commonly uses structured instruments such as VSS and POSAS~\cite{baryza1995vancouver,draaijers2004posas}. Their named dimensions make clinical reasoning explicit, but several items require palpation or patient report and cannot be recovered from an ordinary photograph. Earlier computational pipelines translated observable color, texture, and morphology into handcrafted descriptors. Such features are inspectable but laborious to design and brittle under changes in illumination or acquisition.

Deep networks instead learn image representations and have achieved strong results in dermatology and medical imaging~\cite{esteva2017dermatologist,liu2020derm,litjens2017survey}. Recent dermatology foundation models and ontology-aligned vision--language resources extend this trend across broader tasks~\cite{yan2025panderm,yan2025derm1m}. Their effectiveness nevertheless depends on task-relevant data and validation under the intended deployment shift. For KD--HS differentiation specifically, a pilot that repeatedly queried chat-based models with scar images reported low diagnostic accuracy and concluded that direct use was not clinically ready~\cite{shiraishi2024potential}. We evaluate supervised networks and frozen visual encoders under patient-level, leave-one-hospital-out testing.

\subsection{Language and Vision-Language Models in Medicine}
Medical LLMs encode clinical knowledge, while vision--language models extend it to biomedical images and report generation~\cite{singhal2023large,nori2023capabilities,zhou2024pre,li2023llava,rao2025multimodal,sellergren2026medgemma15}. These systems map an image to a prediction, answer, or explanation; a recent evaluation further shows that multimodal medical predictions can rely strongly on accompanying text~\cite{buckley2026multimodal}. Hosted inference requires transmitting the image, whereas local deployment still couples the decision to a large, opaque model. ScaFE instead asks the LLM to produce an inspectable program without seeing a clinical image; the program then executes deterministically inside the hospital without an LLM at inference time.

\subsection{Knowledge-Guided Structured Representations}
Predictive performance depends on data representation and on avoiding common evaluation pitfalls~\cite{bengio2013representation,domingos2012few}. Concept bottleneck models learn named intermediate concepts~\cite{koh2020concept}, while knowledge-guided networks inject clinical priors into model training~\cite{xie2019knowledge}. Neurosymbolic work integrates neural learning with symbolic knowledge~\cite{garcez2023neural}; separately, prototype methods organize tabular representations around explicit global prototypes~\cite{ye2024ptarl}. ScaFE differs in where the representation comes from: it does not learn a concept layer from the same small image cohort or rely on a manually fixed feature set. A web-enabled LLM converts source-traceable clinical concepts into executable measurements, and held-out aggregate feedback iteratively improves the program. This separation provides data-efficient learning, local image processing, and feature-level auditability within one pipeline.

%% file: sections/preliminary.tex
\section{Problem Formulation}
\label{sec:problem}

\paragraph{Scar image classification.}
Let $\mathcal{X}\subset\mathbb{R}^{H\times W\times 3}$ denote the space of RGB scar images, where $H,W\in\mathbb{N}$ are the image height and width, and let $\mathcal{Y}=\{1,\ldots,C\}$ denote $C$ diagnostic categories. We are given a labeled development set and a disjoint test set,
\begin{equation}
    \mathcal{D}_{\mathrm{train}}
    =\{(I_i,y_i)\}_{i=1}^{N},
    \qquad
    \mathcal{D}_{\mathrm{test}}
    =\{(I_i,y_i)\}_{i=N+1}^{N+N_{\mathrm{te}}},
\end{equation}
where $I_i\in\mathcal{X}$ and $y_i\in\mathcal{Y}$. The objective is to learn a predictor from $\mathcal{D}_{\mathrm{train}}$ that generalizes to $\mathcal{D}_{\mathrm{test}}$. The test set is excluded from feature construction, classifier fitting, and model selection.

\paragraph{Program-based representation.}
Instead of learning an image representation and a decision function jointly, we decompose the predictor into an executable feature program $g$ and a classifier $h_g:\mathbb{R}^{K_g}\rightarrow\mathcal{Y}$:
\begin{equation}
    g:\mathcal{X}\rightarrow\mathbb{R}^{K_g},
    \qquad
    f_g(I)=h_g(g(I)).
    \label{eq:decomposition}
\end{equation}
The program-dependent dimension $K_g$ allows different candidates to encode different sets of visual measurements. We call $g$ \emph{executable} if it terminates under the prescribed resource limit and returns a deterministic, finite, fixed-length numeric vector for every valid input. Let $\mathcal{G}_{\mathrm{exec}}$ denote the set of executable candidates.

\paragraph{Validation-guided program search.}
We partition the development data into fitting and validation subsets,
\begin{equation}
    \mathcal{D}_{\mathrm{train}}
    =\mathcal{D}_{\mathrm{fit}}\mathbin{\dot\cup}
    \mathcal{D}_{\mathrm{val}}.
    \label{eq:development-split}
\end{equation}
Let $\operatorname{TrainRF}$ denote the fixed RF training procedure. For a candidate $g$, the downstream classifier is fitted only on the structured training pairs
\begin{equation}
    h_g=\operatorname{TrainRF}
    \big(\{(g(I_i),y_i):(I_i,y_i)\in\mathcal{D}_{\mathrm{fit}}\}\big).
\end{equation}
Given a scalar validation metric $\mathcal{M}$, our goal is to search for
\begin{equation}
    g^*
    =\arg\max_{g\in\mathcal{G}_{\mathrm{exec}}}
    \mathcal{M}(h_g\circ g;\mathcal{D}_{\mathrm{val}}).
    \label{eq:program-objective}
\end{equation}
ScaFE performs this search by asking an LLM to generate and revise feature programs. The LLM observes online clinical evidence, candidate source code, and aggregate validation feedback, but not raw images or patient-level records.

%% file: sections/method.tex
\section{The ScaFE Framework}
\label{sec:method}

\begin{figure*}[t]
    \centering
    \includegraphics[width=0.86\textwidth]{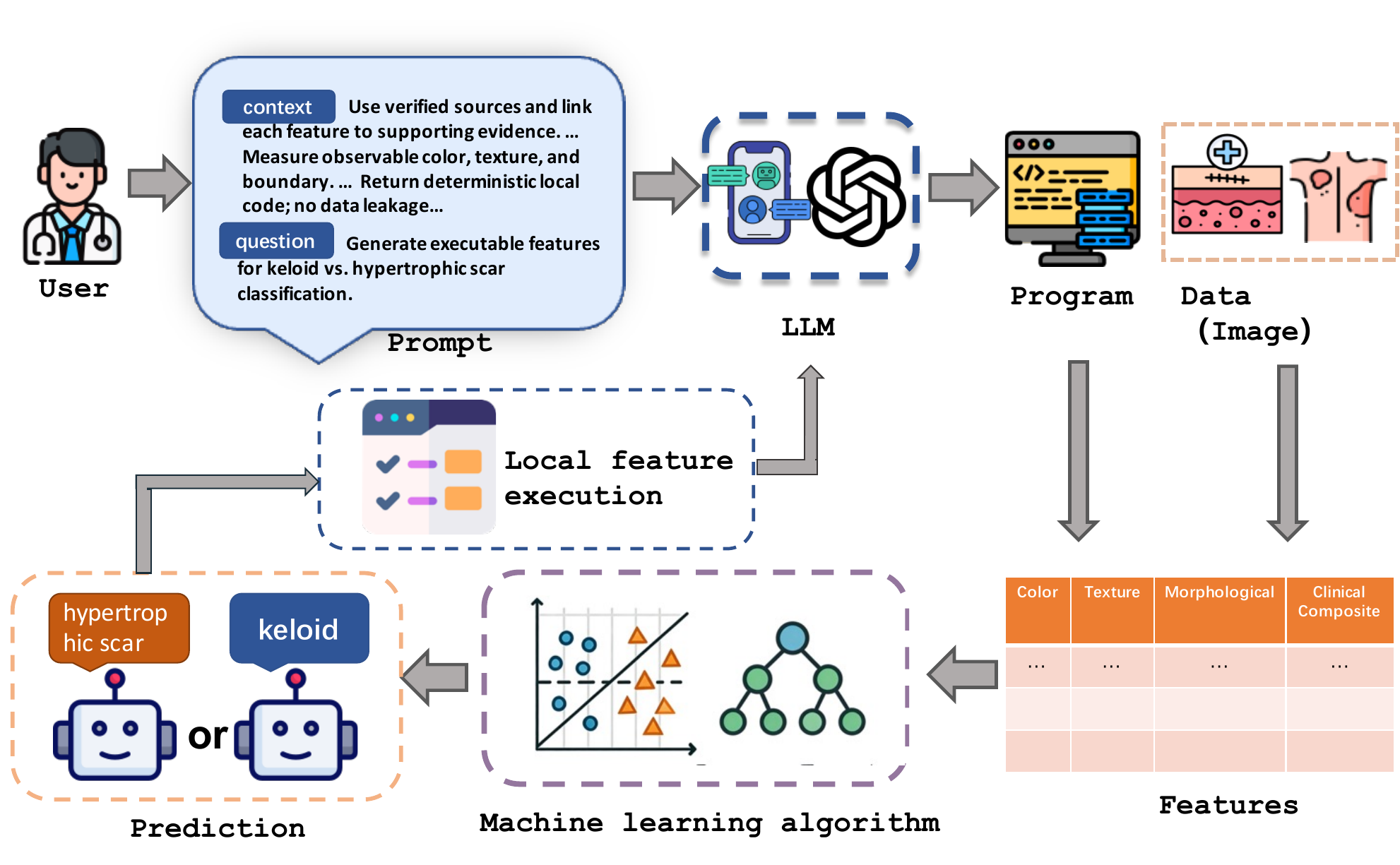}
    \caption{Overview of ScaFE. An LLM turns source-traceable clinical concepts into deterministic feature programs that process local images into interpretable features. A fixed RF evaluates candidates, and only aggregate diagnostics return for refinement. The selected program and RF are frozen for prediction; raw images never enter the LLM context.}
    \label{fig:framework}
\end{figure*}

ScaFE converts online clinical evidence into executable image features through three stages: web-grounded candidate generation, validation-guided local refinement, and final model construction with a Random Forest (RF). Figure~\ref{fig:framework} summarizes the separation between LLM knowledge transfer, local feature execution, and downstream prediction.

\subsection{Overview}

ScaFE runs $T$ rounds. At round $t\in\{1,\ldots,T\}$, the LLM retrieves public clinical evidence and proposes $M$ feature programs. Each program executes locally, and a fixed RF trained on $\mathcal{D}_{\mathrm{fit}}$ evaluates it on $\mathcal{D}_{\mathrm{val}}$. Only aggregate diagnostics return to the LLM to guide the next repair. After round $T$, ScaFE selects one program, refits the RF on all development data, and evaluates the frozen pipeline on $\mathcal{D}_{\mathrm{test}}$.

The LLM may search public literature, but generated code has no network, file-system, label, filename, or metadata access. Raw clinical images remain local throughout the search.

\subsection{Web-Grounded Feature Program Generation}

\paragraph{Autonomous literature search.}
Let $\mathcal{L}_{\Theta}$ denote a fixed LLM configuration indexed by $\Theta$, and let $\mathcal{B}$ denote an online search operator that maps a set of queries to retrieved evidence records. ScaFE does not provide a predetermined reference collection. Instead, the prompt describes the diagnostic task and permits $\mathcal{L}_{\Theta}$ to formulate its own query set $\mathcal{Q}_t$. At round $t$, the retrieved evidence set is
\begin{equation}
    \mathcal{R}_t=\mathcal{B}(\mathcal{Q}_t).
    \label{eq:web-retrieval}
\end{equation}
The search prioritizes peer-reviewed articles, official guidelines, and primary descriptions of criteria such as VSS and POSAS~\cite{baryza1995vancouver,draaijers2004posas}. Each evidence record stores bibliographic identifiers, access time, a supporting passage, and the supported clinical concept. A record enters $\mathcal{R}_t$ only if its identifier resolves and its passage is recoverable. Queries, verification outcomes, and records are archived with the code. Listing~1 in Section~B of the Supplementary Document gives the exact contract.

\paragraph{Candidate program generation.}
Let $\mathcal{P}_t$ denote the prompt used at round $t$, and let $\mathcal{S}_{t-1}$ denote the collection of aggregate feedback records from the previous round. The first round uses the generation prompt $\mathcal{P}_{\mathrm{gen}}$ and evidence $\mathcal{R}_1$; later rounds additionally receive the previous programs and feedback. We write this process as
\begin{equation}
    \mathcal{G}_t
    =\mathcal{L}_{\Theta}
    \big(\mathcal{P}_t,\mathcal{R}_t,\mathcal{G}_{t-1},\mathcal{S}_{t-1}\big),
    \qquad
    \mathcal{G}_t=\{g_t^{(m)}\}_{m=1}^{M},
    \label{eq:candidate-generation}
\end{equation}
where $\mathcal{P}_1=\mathcal{P}_{\mathrm{gen}}$ and $\mathcal{G}_0=\mathcal{S}_0=\varnothing$. Each candidate implements
\begin{equation}
    g_t^{(m)}(I)
    =[\phi_{t,m,1}(I),\ldots,\phi_{t,m,K_{t,m}}(I)]^\top
    \in\mathbb{R}^{K_{t,m}}.
    \label{eq:feature-program}
\end{equation}
Here $K_{t,m}=K_{g_t^{(m)}}$ is the candidate's output dimension and $\phi_{t,m,j}$ is its $j$th scalar feature. Alongside the code, the LLM returns the name, definition, clinical rationale, and online source record for every output dimension. Listing~2 in Section~B of the Supplementary Document gives the complete generation template and output contract.

\subsection{Validation-Guided Iterative Refinement}

\paragraph{Local execution check.}
Each program is parsed and run under fixed time and memory budgets. Valid code uses only approved libraries and returns a deterministic, finite, fixed-length vector. Failures yield a sanitized category and message; the triggering image and intermediate values remain local. Before evaluation, dimensions lacking a valid source record and recoverable passage are removed, and candidates with none are discarded. Thus every downstream feature passes the evidence gate. These checks audit reliability and traceability; held-out evaluation measures utility.

\paragraph{Candidate evaluation.}
For $A\in\{\mathrm{fit},\mathrm{val},\mathrm{train}\}$, let $N_A=|\mathcal{D}_A|$. Applying an executable program $g$ row-wise gives
\begin{equation}
\begin{aligned}
    X_A^g&=[g(I_i)^\top]_{(I_i,y_i)\in\mathcal{D}_A}
    \in\mathbb{R}^{N_A\times K_g},\\
    Y_A&=[y_i]_{(I_i,y_i)\in\mathcal{D}_A}
    \in\mathcal{Y}^{N_A}.
\end{aligned}
\end{equation}
We fit $h_g$ on $(X_{\mathrm{fit}}^{g},Y_{\mathrm{fit}})$ and predict the validation samples. The class-wise results are summarized by
\begin{equation}
    C_{ab}^{g}
    =\sum_{(I_i,y_i)\in\mathcal{D}_{\mathrm{val}}}
    \mathbf{1}\!\left[y_i=a\wedge h_g(g(I_i))=b\right],
    \label{eq:confusion}
\end{equation}
where $\mathbf{1}[\cdot]$ is the indicator function and, for $a,b\in\mathcal{Y}$, $C_{ab}^{g}$ counts validation examples from class $a$ predicted as class $b$. The scalar comparison metric in Eq.~\eqref{eq:program-objective} is balanced accuracy,
\begin{equation}
    \operatorname{BAcc}_{\mathrm{val}}(g)
    =\frac{1}{C}\sum_{c=1}^{C}
    \frac{C_{cc}^{g}}{\sum_{b=1}^{C}C_{cb}^{g}}.
    \label{eq:balanced-accuracy}
\end{equation}

\paragraph{Aggregate feature feedback.}
Classification counts alone do not indicate whether a program produces unusable or ignored features. ScaFE therefore also reports the feature-matrix shapes, the fractions of non-finite and constant dimensions, and global feature importance. We compute SHAP values~\cite{NIPS2017_8a20a862} for the validation predictions and aggregate feature $j$ as
\begin{equation}
    \bar{\psi}_{j}^{g}
    =\frac{1}{N_{\mathrm{val}}C}
    \sum_{i=1}^{N_{\mathrm{val}}}\sum_{c=1}^{C}
    |\psi_{ijc}^{g}|,
    \label{eq:mean-shap}
\end{equation}
where $\psi_{ijc}^{g}$ is the attribution of feature $j\in\{1,\ldots,K_g\}$ to class $c$ for validation sample $i$. Let $C^g=(C_{ab}^g)_{a,b=1}^{C}$, $\bar{\boldsymbol{\psi}}^g=(\bar\psi_j^g)_{j=1}^{K_g}$, and let $r_{\mathrm{invalid}}$ and $r_{\mathrm{constant}}$ be the fractions of invalid and constant output dimensions. The candidate-level feedback record is
\begin{equation}
\begin{split}
    S_t(g)=\{&
    \text{status},\text{error},C^g,\operatorname{BAcc}_{\mathrm{val}}(g),\\
    &\operatorname{shape}(X_{\mathrm{fit}}^g),
    \operatorname{shape}(X_{\mathrm{val}}^g),\\
    &
    r_{\mathrm{invalid}},r_{\mathrm{constant}},
    \bar{\boldsymbol{\psi}}^g\}.
\end{split}
    \label{eq:feedback}
\end{equation}
The round-level collection is $\mathcal{S}_t=\{S_t(g):g\in\mathcal{G}_t\}$. Only these aggregate records are exposed to the LLM; patient-level features, predictions, and SHAP values remain local. The machine-readable fields are reproduced in Listing~4 of the Supplementary Document.

\paragraph{Program refinement.}
The LLM repairs execution failures, preserves supported features with useful validation contributions, and revises weak features; it may search again for a missing clinical concept. Listing~3 of the Supplementary Document gives the exact template. We fix $M$, $T$, the split in Eq.~\eqref{eq:development-split}, and the RF before search. Because $\mathcal{D}_{\mathrm{val}}$ is queried repeatedly, it is an optimization set; only the sealed test set estimates generalization.

\subsection{Random Forest as the Downstream Learner}

\paragraph{Rationale.}
A feature program produces a low-dimensional table of heterogeneous color, texture, and morphology measurements rather than a spatial tensor. RF accommodates mixed scales without standardization, captures nonlinear thresholds and interactions, and reduces tree variance through bootstrap aggregation and feature subsampling~\cite{breiman2001random}.

\paragraph{Role in ScaFE.}
ScaFE fixes one RF configuration across candidates and rounds, so validation changes primarily reflect the representation rather than classifier tuning. RF serves as both search-time evaluator and final learner. A bounded search budget and sealed test set, not the RF itself, separate search from evaluation.

\subsection{Final Model Construction and Inference}

Let $\mathcal{G}_{T}^{\mathrm{exec}}$ denote the executable candidates in the last round. We select
\begin{equation}
    g^*
    =\arg\max_{g\in\mathcal{G}_{T}^{\mathrm{exec}}}
    \operatorname{BAcc}_{\mathrm{val}}(g),
    \label{eq:final-program}
\end{equation}
breaking exact ties by the smaller feature dimension. We then extract features from the complete development set and refit the RF:
\begin{equation}
    h^*=\operatorname{TrainRF}
    (X_{\mathrm{train}}^{g^*},Y_{\mathrm{train}}).
    \label{eq:final-classifier}
\end{equation}
For a new image $I$, ScaFE predicts $\hat{y}=h^*(g^*(I))$. Both $g^*$ and $h^*$ are frozen before evaluating $\mathcal{D}_{\mathrm{test}}$.

Algorithm~\ref{alg:scafe} summarizes the complete procedure.

\begin{algorithm}[t]
\caption{Validation-Guided ScaFE}
\label{alg:scafe}
\footnotesize
\begin{algorithmic}[1]
\Require $\mathcal{D}_{\mathrm{train}}$, sealed $\mathcal{D}_{\mathrm{test}}$, LLM $\mathcal{L}_{\Theta}$, web tool $\mathcal{B}$, rounds $T$, candidates $M$
\Ensure Feature program $g^*$ and RF classifier $h^*$
\State Split $\mathcal{D}_{\mathrm{train}}$ into $\mathcal{D}_{\mathrm{fit}},\mathcal{D}_{\mathrm{val}}$; set $\mathcal{G}_0,\mathcal{S}_0\gets\varnothing$
\For{$t=1$ to $T$}
    \State $\mathcal{Q}_t,\mathcal{R}_t\gets\Call{Search}{\mathcal{L}_{\Theta},\mathcal{B},\mathcal{P}_t,\mathcal{S}_{t-1}}$
    \State $\mathcal{G}_t\gets\Call{Generate}{\mathcal{P}_t,\mathcal{R}_t,\mathcal{G}_{t-1},\mathcal{S}_{t-1},M}$
    \State $\mathcal{S}_t\gets\{\Call{Evaluate}{g,\mathcal{D}_{\mathrm{fit}},\mathcal{D}_{\mathrm{val}}}:g\in\mathcal{G}_t\}$
\EndFor
\State Select $g^*$ from $\mathcal{G}_{T}^{\mathrm{exec}}$ by validation BAcc
\State Fit $h^*\gets\operatorname{TrainRF}(X_{\mathrm{train}}^{g^*},Y_{\mathrm{train}})$; evaluate once on $\mathcal{D}_{\mathrm{test}}$
\State \Return $g^*,h^*$
\end{algorithmic}
\end{algorithm}

%% file: sections/experiment.tex
\providecommand{\resultvalue}[1]{#1}

\section{Experiments}
\label{sec:experiments}

Our evaluation is organized around four questions. \textbf{RQ1} asks whether
ScaFE generalizes to an unseen clinical site better than classical features,
supervised transfer learning, visual foundation models, and direct VLM
inference. \textbf{RQ2} identifies which parts of the iterative search account
for any gain and whether RF is an appropriate downstream learner. \textbf{RQ3}
examines data efficiency and robustness to acquisition shifts. \textbf{RQ4}
tests whether the generated programs are executable, evidence-grounded,
stable, interpretable, and computationally practical.

\subsection{Experimental Protocol}

\paragraph{Multi-institutional cohort.}
We study binary classification of keloids (KD) and hypertrophic scars (HS) on
600 de-identified clinical photographs collected at three hospitals. Each site
contributes 200 images, with 100 KD and 100 HS cases.
The cohort contains 600 unique patients (200 per site; one image per patient).
Duplicate and near-duplicate images are removed before any split is formed.
Reference diagnoses are taken from the clinical record and independently
verified by two board-certified plastic surgeons with 8 and 11 years of
experience; disagreements are resolved by a third specialist with 15 years of
experience. Inter-rater agreement before adjudication is Cohen's $\kappa=0.86$.
Images were acquired from January 2020 to December 2025 using smartphones
(72\%) and DSLR cameras (28\%). Inclusion requires an adult patient, a
record-confirmed KD or HS diagnosis, and a focused pre-treatment photograph;
images with an obscured lesion, postoperative dressing, or unresolved duplicate
are excluded. The retrospective, de-identified study was approved by the ethics
committees of the participating institutions with a waiver of additional
consent; identifying approval numbers are omitted for blind review. The
clinicians establish reference labels only and do not design features, inspect
search feedback, or participate in ScaFE. Table~1 in Section~A.1 of the
Supplementary Document gives the per-site composition and further reports
exclusions, demographic coverage, missingness, and patient/duplicate leakage
checks. None of these attributes is used as input.

\paragraph{Nested cross-site evaluation.}
We use three leave-one-site-out folds. In each fold, one hospital (200 images)
is sealed as the test set and the other two hospitals (400 images) form the
development set. The latter is divided patient-wise into 80\% fitting and 20\%
validation data, stratified by site and diagnosis. The fitting subset trains
classifiers, whereas the validation subset supplies all program-refinement,
hyperparameter-selection, and early-stopping signals. ScaFE is rerun from
scratch within every outer fold. The held-out site is opened only after its
feature program, classifier, and all thresholds have been frozen. Every method
uses the same partitions; neither site identifiers nor test outcomes are
available to generated code or to the LLM.

\paragraph{Metrics and statistical analysis.}
The primary endpoint is site-macro balanced accuracy (BAcc): BAcc is computed
separately on Sites A--C and then averaged without weighting by site size. We
also report macro-F1, AUROC, sensitivity, specificity, Brier score, and expected
calibration error. Stochastic procedures are repeated with five prespecified
seeds. We report their mean and a 95\% confidence interval obtained from 10,000
paired, patient-level bootstrap resamples within each site. Pairwise
improvements over all baselines use the same resamples and Holm correction.
Site-level results and calibration curves are retained rather than reporting
only a single cross-site average. Because there are only three hospitals, these
intervals quantify patient sampling within the observed sites, not uncertainty
over the wider population of hospitals.

\paragraph{Implementation details.}
Images are stripped of metadata, orientation-corrected, and resized with
aspect-ratio-preserving padding to $512\!\times\!512$ for ScaFE; neural models
use their native input resolution. No manual crop, lesion mask, or clinical
metadata is supplied. We generate $M=4$ programs in each of $T=3$ rounds with
temperature 0.2 using \texttt{gpt-4.1-2025-04-14} and web search (accessed July
15--20, 2026). Every candidate uses the same RF: 500 trees, depth 8, minimum
leaf size 3, $\sqrt{K_g}$ features per split for a $K_g$-dimensional program,
class balancing, and a fixed seed.
Table~2 in Section~A.2 of the Supplementary Document lists the complete
selection budgets, software, model revisions, augmentation policy, and archived
run artifacts.

\subsection{Baselines and Controls}

We compare ScaFE with complementary, non-strawman alternatives.
\textbf{Clinical handcrafted+RF} uses prespecified color histograms, GLCM/LBP
texture, and threshold-based morphology motivated by VSS and POSAS, without an
LLM. \textbf{ResNet-18}~\cite{he2016deep} and
\textbf{EfficientNet-B0}~\cite{tan2019efficientnet} are initialized from
ImageNet weights and fine-tuned with validation-based early stopping.
Frozen \textbf{DINOv3}~\cite{simeoni2025dinov3}, \textbf{Derm
Foundation}~\cite{kiraly2024haidef}, and \textbf{BiomedCLIP}~\cite{zhang2025biomedclip}
encoders use $\ell_2$-regularized linear probes. \textbf{Local VLM-direct}
prompts MedGemma-1.5-4B-IT~\cite{sellergren2026medgemma15} inside the hospital
environment and normalizes the sequence likelihoods of the two allowed labels.
Each learned baseline receives at most $MT=12$ validation configurations.

Two negative controls test arbitrary frozen ResNet features and 1,000 within-site
development-label permutations with preserved class counts. Both use held-out
labels only for evaluation and cannot encode sample identity.

\subsection{Cross-Site Generalization (RQ1)}

Table~\ref{tab:main-results} is the primary comparison. The three site columns
make a result auditable: a high average cannot conceal failure at one hospital.
Tables~6 and 7 in Section~A.4 of the Supplementary Document supply the complete
secondary metrics, confidence intervals, and leakage controls.

\begin{table}[t]
\centering
\caption{Patient-level leave-one-site-out performance. Site A--C columns report
held-out BAcc (\%); Avg. is their unweighted mean. Macro-F1, AUROC, calibration,
and confidence intervals appear in Section~A.4 of the Supplementary Document.}
\label{tab:main-results}
\scriptsize
\setlength{\tabcolsep}{4pt}
\begin{tabular}{lrrrr}
\toprule
Method & Site A & Site B & Site C & Avg. \\
\midrule
Handcrafted + RF        & \resultvalue{62.0} & \resultvalue{60.5} & \resultvalue{61.5} & \resultvalue{61.3} \\
ResNet-18               & \resultvalue{65.0} & \resultvalue{63.0} & \resultvalue{64.5} & \resultvalue{64.2} \\
EfficientNet-B0         & \resultvalue{66.0} & \resultvalue{64.0} & \resultvalue{65.5} & \resultvalue{65.2} \\
DINOv3 + linear probe   & \resultvalue{70.0} & \resultvalue{67.5} & \resultvalue{68.5} & \resultvalue{68.7} \\
Derm Foundation + probe& \resultvalue{71.5} & \resultvalue{68.5} & \resultvalue{69.0} & \resultvalue{69.7} \\
BiomedCLIP + linear probe& \resultvalue{\underline{72.5}} & \resultvalue{\underline{69.5}} & \resultvalue{\underline{71.0}} & \resultvalue{\underline{71.0}} \\
Local VLM-direct        & \resultvalue{67.0} & \resultvalue{65.5} & \resultvalue{66.0} & \resultvalue{66.2} \\
\midrule
ScaFE (ours)            & \resultvalue{\textbf{82.5}} & \resultvalue{\textbf{80.0}} & \resultvalue{\textbf{80.5}} & \resultvalue{\textbf{81.0}} \\
\bottomrule
\end{tabular}
\end{table}

\paragraph{Results.}
ScaFE obtains 81.0\% site-macro BAcc, 10.0 points above the strongest baseline,
BiomedCLIP (71.0\%). The margins are 10.0, 10.5, and 9.5 points on Sites A--C,
respectively, so the average does not conceal a site-specific failure. The
paired patient-level bootstrap gives a 95\% CI of 7.2--12.8 points for the
improvement, with Holm-adjusted $p<0.001$.
ScaFE also reaches 80.7\% macro-F1 and 87.9\% AUROC, while reducing ECE to
0.043. The random-weight encoder and label-permutation controls remain near
chance at 51.2\% and 50.0\% BAcc, respectively, arguing against an obvious
sample-identity or split-leakage explanation for the gain.

\subsection{What Produces the Gain? (RQ2)}

\begin{figure*}[t]
\centering
\includegraphics[width=0.96\textwidth]{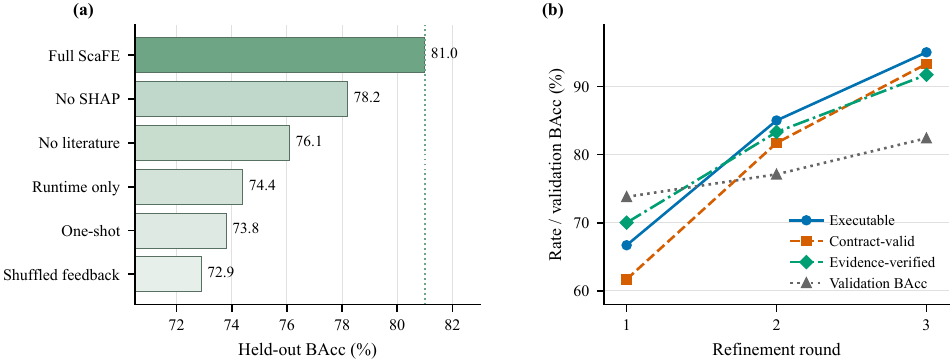}
\caption{Search-loop analysis. (a) Held-out BAcc for ScaFE and its ablations.
(b) Executability, contract validity, evidence verification, and validation
BAcc across refinement rounds. Test BAcc is computed only after freezing the
pipeline; exact values appear in Section~A.3 of the Supplementary Document.}
\label{fig:search-diagnostics}
\end{figure*}

\paragraph{Search-mechanism ablations.}
Figure~\ref{fig:search-diagnostics}(a) separates clinical grounding, mechanical code
repair, prediction-level feedback, and feature-level attribution. The one-shot
variant generates $M$ candidates only at $t=1$. ``Runtime feedback only''
passes syntax and contract failures but withholds confusion counts, BAcc, and
SHAP. ``No SHAP'' retains confusion counts and BAcc. In the shuffled-feedback
control, aggregate records are randomly reassigned among candidates within a
round, preserving their marginal values but destroying their semantic link to
the program being revised.

\paragraph{Downstream learner.}
To test rather than assume the RF choice, we freeze each selected feature
program and replace only its classifier. Logistic
regression and RBF-SVM test linear and smooth nonlinear boundaries, a single
decision tree exposes the variance of unbagged trees, and
XGBoost~\cite{chen2016xgboost} provides a strong boosting comparator. All
hyperparameters are selected on the inner validation set under the same budget;
Table~5 in Section~A.3 of the Supplementary Document gives all five classifier
results.

\paragraph{Results.}
Iterative refinement raises the executable-program rate from 66.7\% to 95.0\%
and BAcc from 73.8\% to 81.0\%. Removing online literature, prediction feedback,
or SHAP loses 4.9, 6.6, and 2.8 points, respectively. Shuffling candidate
feedback reduces BAcc to 72.9\%, slightly below the one-shot variant, while
retaining a high execution rate; this pattern isolates the value of
semantically aligned feedback from that of additional LLM calls. RF has the
highest BAcc and, across the same five synthesis trajectories, the lowest
pipeline SD. XGBoost is close in mean BAcc (80.4\%) but less stable (1.9 versus
1.1 points).

\subsection{Data Efficiency and Robustness (RQ3)}

\begin{figure}[t]
\centering
\includegraphics[width=0.98\columnwidth]{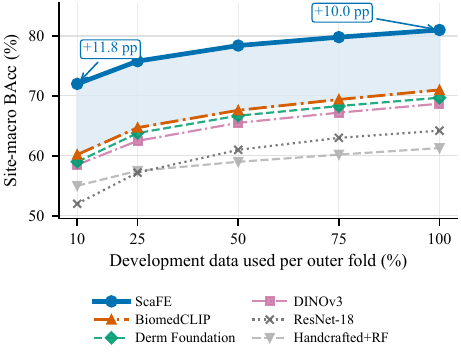}
\caption{Data efficiency under stratified subsampling. Fractions
10/25/50/75/100 use approximately 40/100/200/300/400 development images per
fold; shading marks ScaFE's gain over the strongest baseline, BiomedCLIP.}
\label{fig:data-efficiency}
\end{figure}

We subsample 10\%, 25\%, 50\%, and 75\% of each development set while preserving
site, class, and patient groups. Crucially, the entire ScaFE search is rerun
inside each subsample; discovering a program with all 400 development images
and only refitting its classifier on fewer images would leak information into
the low-data curve. Figure~\ref{fig:data-efficiency} reports site-macro BAcc.

We additionally evaluate frozen models under brightness and contrast shifts,
JPEG compression, and mild Gaussian blur, with perturbation strengths fixed
before test evaluation. Section~A.5 of the Supplementary Document reports this
acquisition-shift audit without elevating it to a separate robustness claim.
Where demographic and body-site metadata are sufficiently complete, we audit
coverage and missingness without using those attributes as inputs.

\paragraph{Results.}
ScaFE obtains BAcc values of 72.0\%, 75.8\%, 78.4\%, 79.8\%, and 81.0\% as the
development fraction increases. Its lead over the strongest baseline is largest
at 10\% (11.8 points) and remains 10.0 points with all development data. Under
the prespecified acquisition shifts, every model degrades; ScaFE retains the
highest absolute BAcc, but its relative degradation is comparable to the
baselines and we therefore make no separate robustness claim.

\subsection{Reliability, Interpretability, and Cost (RQ4)}

RQ4 audits execution, source traceability, feature faithfulness, and
cross-trajectory stability. Domain experts assess the retained source--feature
mappings and visual proxies and confirm their clinical reasonableness.
Supplementary Section~A.6 reports this audit, second-LLM replication, and cost.

\paragraph{Results.}
Executability rises from \resultvalue{66.7\% to 95.0\%}; candidate evidence
passes at \resultvalue{91.7\%}, while final coverage is \resultvalue{100\%}.
Top-three SHAP deletion loses \resultvalue{8.6 points} versus \resultvalue{2.1}
for random deletion (\resultvalue{$p<0.001$}), and trajectories agree on
\resultvalue{$91.4\pm1.8$\%} of predictions. Search takes \resultvalue{12 calls
and 18.6 minutes per fold}, whereas local extraction takes
\resultvalue{34.7 ms/image}. A second LLM remains comparable at
\resultvalue{$79.6\pm1.5$\% BAcc} (Supplementary Section~A.6).

%% file: sections/conclusion.tex
\section{Conclusion}

ScaFE converts LLM clinical knowledge into locally executed, evidence-grounded
feature programs. On 600 photographs from three hospitals, it achieves 81.0\%
site-macro BAcc---10.0 points above the strongest baseline and 11.8 points ahead
with only 10\% of the development data. Refinement raises executability from
66.7\% to 95.0\% and the candidate evidence-pass rate from 70.0\% to 91.7\%;
final filtering ensures 100\% coverage. Ablations attribute the gains to
grounding, semantic feedback, and feature importance.
ScaFE thus separates knowledge transfer from clinical-data access while
retaining a data-efficient, auditable, and reproducible prediction pipeline.

\section{Limitations}

ScaFE supports, not replaces, diagnosis. This retrospective two-class, three-hospital
study requires prospective clinician validation and omits nonvisual findings.
Deployment requires pinned artifacts, sandboxing, and governance.

%% file: aaai2027.bib
@article{bayat2003skin,
  author = "Bayat, Ardeshir and McGrouther, D. Angus and Ferguson, Mark W. J.",
  title = "Skin scarring",
  journal = "BMJ",
  volume = "326",
  number = "7380",
  pages = "88--92",
  year = "2003"
}

@article{berman2017keloids,
  author = "Berman, Brian and Maderal, Andrea and Raphael, Briana",
  title = "Keloids and hypertrophic scars: pathophysiology, classification, and treatment",
  journal = "Dermatologic Surgery",
  volume = "43",
  pages = "S3--S18",
  year = "2017"
}

@article{baryza1995vancouver,
  author = "Baryza, Mary Jo and Baryza, Gregory A.",
  title = "The {Vancouver Scar Scale}: an administration tool and its interrater reliability",
  journal = "Journal of Burn Care \& Rehabilitation",
  volume = "16",
  number = "5",
  pages = "535--538",
  year = "1995",
  doi = "10.1097/00004630-199509000-00013"
}

@article{draaijers2004posas,
  author = "Draaijers, Lieneke J. and Tempelman, Fenike R. H. and Botman, Yvonne A. M. and Tuinebreijer, Wim E. and Middelkoop, Esther and Kreis, Robert W. and van Zuijlen, Paul P. M.",
  title = "The {Patient and Observer Scar Assessment Scale}: a reliable and feasible tool for scar evaluation",
  journal = "Plastic and Reconstructive Surgery",
  volume = "113",
  number = "7",
  pages = "1960--1965",
  year = "2004",
  doi = "10.1097/01.PRS.0000122207.28773.56"
}

@article{esteva2017dermatologist,
  author = "Esteva, Andre and Kuprel, Brett and Novoa, Roberto A. and Ko, Justin and Swetter, Susan M. and Blau, Helen M. and Thrun, Sebastian",
  title = "Dermatologist-level classification of skin cancer with deep neural networks",
  journal = "Nature",
  volume = "542",
  number = "7639",
  pages = "115--118",
  year = "2017",
  doi = "10.1038/nature21056"
}

@inproceedings{he2016deep,
  author = "He, Kaiming and Zhang, Xiangyu and Ren, Shaoqing and Sun, Jian",
  title = "Deep residual learning for image recognition",
  booktitle = "Proceedings of the IEEE Conference on Computer Vision and Pattern Recognition",
  pages = "770--778",
  year = "2016"
}

@inproceedings{tan2019efficientnet,
  author = "Tan, Mingxing and Le, Quoc V.",
  title = "{EfficientNet}: Rethinking model scaling for convolutional neural networks",
  booktitle = "International Conference on Machine Learning",
  pages = "6105--6114",
  year = "2019"
}

@article{singhal2023large,
  author = "Singhal, Karan and Azizi, Shekoofeh and Tu, Tao and Mahdavi, S. Sara and Wei, Jason and Chung, Hyung Won and Scales, Nathan and Tanwani, Ajay and Cole-Lewis, Heather and Pfohl, Stephen and Payne, Perry and Seneviratne, Martin and Gamble, Paul and Kelly, Chris and Babiker, Abubakr and Sch{\"a}rli, Nathanael and Chowdhery, Aakanksha and Mansfield, Philip and Demner-Fushman, Dina and Ag{\"u}era y Arcas, Blaise and Webster, Dale and Corrado, Greg S. and Matias, Yossi and Chou, Katherine and Gottweis, Juraj and Toma{\v{s}}ev, Nenad and Liu, Yun and Rajkomar, Alvin and Barral, Joelle and Semturs, Christopher and Karthikesalingam, Alan and Natarajan, Vivek",
  title = "Large language models encode clinical knowledge",
  journal = "Nature",
  volume = "620",
  number = "7972",
  pages = "172--180",
  year = "2023",
  doi = "10.1038/s41586-023-06291-2"
}

@article{nori2023capabilities,
  author = "Nori, Harsha and King, Nicholas and McKinney, Scott Mayer and Carignan, Dean and Horvitz, Eric",
  title = "Capabilities of {GPT-4} on medical challenge problems",
  journal = "arXiv preprint arXiv:2303.13375",
  year = "2023"
}

@article{liu2020derm,
  author = "Liu, Yuan and Jain, Ayush and Eng, Clara and Way, David H. and Lee, Kang and Bui, Peggy and Kanada, Kimberly and de Oliveira Marinho, Guilherme and Gallegos, Jessica and Gabriele, Sara and Gupta, Vishakha and Singh, Nalini and Natarajan, Vivek and Hofmann-Wellenhof, Rainer and Corrado, Greg S. and Peng, Lily H. and Webster, Dale R. and Ai, Dennis and Huang, Susan J. and Liu, Yun and Dunn, R. Carter and Coz, David",
  title = "A deep learning system for differential diagnosis of skin diseases",
  journal = "Nature Medicine",
  volume = "26",
  number = "6",
  pages = "900--908",
  year = "2020",
  doi = "10.1038/s41591-020-0842-3"
}

@article{zhou2024pre,
  title={Pre-trained multimodal large language model enhances dermatological diagnosis using {SkinGPT-4}},
  author={Zhou, Juexiao and He, Xiaonan and Sun, Liyuan and Xu, Jiannan and Chen, Xiuying and Chu, Yuetan and Zhou, Longxi and Liao, Xingyu and Zhang, Bin and Afvari, Shawn and Gao, Xin},
  journal={Nature Communications},
  volume={15},
  number={1},
  pages={5649},
  year={2024},
  doi={10.1038/s41467-024-50043-3},
  publisher={Nature Publishing Group UK London}
}

@article{bengio2013representation,
  author = "Bengio, Yoshua and Courville, Aaron and Vincent, Pascal",
  title = "Representation learning: A review and new perspectives",
  journal = "IEEE Transactions on Pattern Analysis and Machine Intelligence",
  volume = "35",
  number = "8",
  pages = "1798--1828",
  year = "2013"
}

@article{domingos2012few,
  author = "Domingos, Pedro",
  title = "A few useful things to know about machine learning",
  journal = "Communications of the ACM",
  volume = "55",
  number = "10",
  pages = "78--87",
  year = "2012"
}

@article{garcez2023neural,
  author = "Garcez, Artur d'Avila and Lamb, Luis C.",
  title = "Neurosymbolic {AI}: The 3rd wave",
  journal = "Artificial Intelligence Review",
  volume = "56",
  pages = "12387--12406",
  year = "2023",
  doi = "10.1007/s10462-023-10448-w"
}

@inproceedings{koh2020concept,
  author = "Koh, Pang Wei and Nguyen, Thao and Tang, Yew Siang and Mussmann, Stephen and Pierson, Emma and Kim, Been and Liang, Percy",
  title = "Concept bottleneck models",
  booktitle = "International Conference on Machine Learning",
  pages = "5338--5348",
  year = "2020"
}

@article{xie2019knowledge,
  author = "Xie, Yutong and Xia, Yong and Zhang, Jianpeng and Song, Yang and Feng, Dagan and Fulham, Michael and Cai, Weidong",
  title = "Knowledge-based collaborative deep learning for benign-malignant lung nodule classification on chest {CT}",
  journal = "IEEE Transactions on Medical Imaging",
  volume = "38",
  number = "4",
  pages = "991--1004",
  year = "2019"
}

@article{li2023llava,
  title={{LLaVA-Med}: Training a Large Language-and-Vision Assistant for Biomedicine in One Day},
  author={Li, Chunyuan and Wong, Cliff and Zhang, Sheng and Usuyama, Naoto and Liu, Haotian and Yang, Jianwei and Naumann, Tristan and Poon, Hoifung and Gao, Jianfeng},
  journal={Advances in Neural Information Processing Systems},
  volume={36},
  pages={28541--28564},
  year={2023},
  doi={10.48550/arXiv.2306.00890}
}

@article{shiraishi2024potential,
  title={The potential of chat-based artificial intelligence models in differentiating between keloid and hypertrophic scars: a pilot study},
  author={Shiraishi, Makoto and Miyamoto, Shimpei and Takeishi, Hakuba and Kurita, Daichi and Furuse, Kiichi and Ohba, Jun and Moriwaki, Yuta and Fujisawa, Kou and Okazaki, Mutsumi},
  journal={Aesthetic Plastic Surgery},
  volume={48},
  number={24},
  pages={5367--5372},
  year={2024},
  doi={10.1007/s00266-024-04380-9},
  publisher={Springer}
}

@article{breiman2001random,
  title = "Random forests",
  author = "Breiman, Leo",
  journal = "Machine learning",
  volume = "45",
  number = "1",
  pages = "5--32",
  year = "2001",
  publisher = "Springer"
}

@article{litjens2017survey,
  title={A survey on deep learning in medical image analysis},
  author={Litjens, Geert and Kooi, Thijs and Bejnordi, Babak Ehteshami and Setio, Arnaud Arindra Adiyoso and Ciompi, Francesco and Ghafoorian, Mohsen and van der Laak, Jeroen A. W. M. and van Ginneken, Bram and S{\'a}nchez, Clara I.},
  journal={Medical image analysis},
  volume={42},
  pages={60--88},
  year={2017},
  doi={10.1016/j.media.2017.07.005},
  publisher={Elsevier}
}

@inproceedings{
ye2024ptarl,
title={{PT}a{RL}: Prototype-based Tabular Representation Learning via Space Calibration},
author={Ye, Hangting and Fan, Wei and Song, Xiaozhuang and Zheng, Shun and Zhao, He and Guo, Dandan and Chang, Yi},
booktitle={The Twelfth International Conference on Learning Representations},
year={2024},
url={https://openreview.net/forum?id=G32oY4Vnm8}
}

@inproceedings{NIPS2017_8a20a862,
  author = {Lundberg, Scott M and Lee, Su-In},
  booktitle = {Advances in Neural Information Processing Systems},
  publisher = {Curran Associates, Inc.},
  title = {A Unified Approach to Interpreting Model Predictions},
  url = {https://proceedings.neurips.cc/paper_files/paper/2017/file/8a20a8621978632d76c43dfd28b67767-Paper.pdf},
  volume = {30},
  year = {2017}
}

@article{zhang2025biomedclip,
  author = {Zhang, Sheng and Xu, Yanbo and Usuyama, Naoto and Xu, Hanwen and Bagga, Jaspreet and Tinn, Robert and Preston, Sam and Rao, Rajesh and Wei, Mu and Valluri, Naveen and Wong, Cliff and Tupini, Andrea and Wang, Yu and Mazzola, Matt and Shukla, Swadheen and Liden, Lars and Gao, Jianfeng and Crabtree, Angela and Piening, Brian and Bifulco, Carlo and Lungren, Matthew P. and Naumann, Tristan and Wang, Sheng and Poon, Hoifung},
  title = {A Multimodal Biomedical Foundation Model Trained from Fifteen Million Image--Text Pairs},
  journal = {NEJM AI},
  volume = {2},
  number = {1},
  year = {2025},
  doi = {10.1056/AIoa2400640}
}

@inproceedings{chen2016xgboost,
  author = {Chen, Tianqi and Guestrin, Carlos},
  title = {{XGBoost}: A Scalable Tree Boosting System},
  booktitle = {Proceedings of the 22nd ACM SIGKDD International Conference on Knowledge Discovery and Data Mining},
  pages = {785--794},
  year = {2016},
  doi = {10.1145/2939672.2939785}
}

@article{simeoni2025dinov3,
  author = {Sim{\'e}oni, Oriane and Vo, Huy V. and Seitzer, Maximilian and Baldassarre, Federico and Oquab, Maxime and Jose, Cijo and Khalidov, Vasil and Szafraniec, Marc and Yi, Seungeun and Ramamonjisoa, Micha{\"e}l and Massa, Francisco and Haziza, Daniel and Wehrstedt, Luca and Wang, Jianyuan and Darcet, Timoth{\'e}e and Moutakanni, Th{\'e}o and Sentana, Leonel and Roberts, Claire and Vedaldi, Andrea and Tolan, Jamie and Brandt, John and Couprie, Camille and Mairal, Julien and J{\'e}gou, Herv{\'e} and Labatut, Patrick and Bojanowski, Piotr},
  title = {{DINOv3}},
  journal = {arXiv preprint arXiv:2508.10104},
  year = {2025},
  doi = {10.48550/arXiv.2508.10104}
}

@article{kiraly2024haidef,
  author = {Kiraly, Atilla P. and Baur, Sebastien and Philbrick, Kenneth and Mahvar, Fereshteh and Yatziv, Liron and Chen, Tiffany and Sterling, Bram and George, Nick and Jamil, Fayaz and Tang, Jing and Bailey, Kai and Ahmed, Faruk and Goel, Akshay and Ward, Abbi and Yang, Lin and Sellergren, Andrew and Matias, Yossi and Hassidim, Avinatan and Shetty, Shravya and Golden, Daniel and Azizi, Shekoofeh and Steiner, David F. and Liu, Yun and Thelin, Tim and Pilgrim, Rory and Kirmizibayrak, Can},
  title = {Health {AI} Developer Foundations},
  journal = {arXiv preprint arXiv:2411.15128},
  year = {2024},
  doi = {10.48550/arXiv.2411.15128}
}

@article{sellergren2026medgemma15,
  author = {Sellergren, Andrew and Gao, Chufan and Mahvar, Fereshteh and Kohlberger, Timo and Jamil, Fayaz and Traverse, Madeleine and Tono, Alberto and Sadjad, Bashir and Yang, Lin and Lau, Charles and Yatziv, Liron and Chen, Tiffany and Sterling, Bram and Philbrick, Kenneth and Tiwari, Richa and Liu, Yun and Jajoo, Madhuram and Sankarapu, Chandrashekar and Vispute, Swapnil and Purandare, Harshad and Mishra, Abhishek Bijay and Schmidgall, Sam and Tu, Tao and Palepu, Anil and Park, Chunjong and Strother, Tim and Thapa, Rahul and Cheng, Yong and Singh, Preeti and Black, Kat and Matias, Yossi and Chou, Katherine and Hassidim, Avinatan and Goel, Kavi and Barral, Joelle and Warkentin, Tris and Shetty, Shravya and Webster, Dale and Virmani, Sunny and Steiner, David F. and Kirmizibayrak, Can and Golden, Daniel},
  title = {{MedGemma} 1.5 Technical Report},
  journal = {arXiv preprint arXiv:2604.05081},
  year = {2026},
  doi = {10.48550/arXiv.2604.05081}
}

@article{yan2025panderm,
  author = {Yan, Siyuan and Yu, Zhen and Primiero, Clare and Vico-Alonso, Cristina and Wang, Zhonghua and Yang, Litao and Tschandl, Philipp and Hu, Ming and Ju, Lie and Tan, Gin and Tang, Vincent and Ng, Aik Beng and Powell, David and Bonnington, Paul and See, Simon and Magnaterra, Elisabetta and Ferguson, Peter and Nguyen, Jennifer and Guitera, Pascale and Banuls, Jose and Janda, Monika and Mar, Victoria and Kittler, Harald and Soyer, H. Peter and Ge, Zongyuan},
  title = {A Multimodal Vision Foundation Model for Clinical Dermatology},
  journal = {Nature Medicine},
  volume = {31},
  number = {8},
  pages = {2691--2702},
  year = {2025},
  doi = {10.1038/s41591-025-03747-y}
}

@inproceedings{yan2025derm1m,
  author = {Yan, Siyuan and Hu, Ming and Jiang, Yiwen and Li, Xieji and Fei, Hao and Tschandl, Philipp and Kittler, Harald and Ge, Zongyuan},
  title = {{Derm1M}: A Million-Scale Vision--Language Dataset Aligned with Clinical Ontology Knowledge for Dermatology},
  booktitle = {Proceedings of the IEEE/CVF International Conference on Computer Vision},
  pages = {12681--12690},
  year = {2025}
}

@article{rao2025multimodal,
  author = {Rao, Vishwanatha M. and Hla, Michael and Moor, Michael and Adithan, Subathra and Kwak, Stephen and Topol, Eric J. and Rajpurkar, Pranav},
  title = {Multimodal Generative {AI} for Medical Image Interpretation},
  journal = {Nature},
  volume = {639},
  number = {8056},
  pages = {888--896},
  year = {2025},
  doi = {10.1038/s41586-025-08675-y}
}

@article{buckley2026multimodal,
  author = {Buckley, Thomas A. and Diao, James A. and Srivastava, Cam N. and Brodeur, Peter G. and Rajpurkar, Pranav and Rodman, Adam and Manrai, Arjun K.},
  title = {Multimodal Foundation Models Exploit Text to Make Medical Image Predictions},
  journal = {Nature Communications},
  volume = {17},
  number = {1},
  pages = {7475},
  year = {2026},
  doi = {10.1038/s41467-026-74207-5}
}
